\RequirePackage[T1]{fontenc}
\documentclass[journal,10pt]{IEEEtran}

\ifdefined\pdfsuppresswarningpagegroup
\fi

\usepackage{amsmath,amssymb}
\usepackage{graphicx}
\usepackage{booktabs}
\usepackage{multirow}
\usepackage{array}
\usepackage{tabularx}
\usepackage[table]{xcolor}
\usepackage[nocompress]{cite}
\usepackage{url}
\usepackage{microtype}
\usepackage{algorithmic}
\usepackage{placeins}
\usepackage{capt-of}
\usepackage[caption=false,font=footnotesize]{subfig}
\usepackage[hidelinks]{hyperref}
\newcounter{algorithm}

\makeatletter
\let\ps@paceoriginaltitle\ps@IEEEtitlepagestyle
\def\ps@pacepreprint{\ps@paceoriginaltitle
  \def\@oddfoot{\normalfont\footnotesize Preprint\hfil}%
  \let\@evenfoot\@oddfoot}
\newcommand{\resultsubsection}{\@startsection{subsection}{2}{\z@}%
  {3.0ex minus 0.5ex}{0.7ex}{\normalfont\normalsize\itshape}}
\newcommand{\compactsection}{\@startsection{section}{1}{\z@}%
  {2.5ex minus 0.5ex}{0.7ex}{\normalfont\normalsize\centering\scshape}}
\makeatother

\definecolor{rowblue}{HTML}{EAF3F1}

\newcolumntype{Y}{>{\centering\arraybackslash}X}
\newcolumntype{L}[1]{>{\raggedright\arraybackslash}p{#1}}
\newcolumntype{C}[1]{>{\centering\arraybackslash}p{#1}}

\title{PACE-Mem: Prefix-Aware Complementary Evidence for Multimodal Agents}

\author{Shidu~Ren, Rong-cheng~Tu, Hang~Zhou, and Xiao~Luo%
\thanks{Code: \protect\url{https://github.com/Shidu-Ren/PACE-Mem}.}%
\thanks{Corresponding author: Rong-cheng~Tu.}%
\thanks{Shidu Ren is with the Edward S. Rogers Sr. Department of Electrical and Computer Engineering, University of Toronto, Toronto, ON M5S 3G4, Canada (e-mail: ryan.ren@mail.utoronto.ca).}%
\thanks{Rong-cheng Tu is with the College of Computing and Data Science, Nanyang Technological University, Singapore 639798 (e-mail: rongcheng.tu@ntu.edu.sg).}%
\thanks{Hang Zhou is with the Department of Statistics and Operations Research and the School of Data and Information Science, The University of North Carolina at Chapel Hill, Chapel Hill, NC 27599 USA (e-mail: hzh@unc.edu).}%
\thanks{Xiao Luo is with the Department of Statistics, University of Wisconsin--Madison, Madison, WI 53706 USA (e-mail: xiao.luo@wisc.edu).}}

\begin{document}
\bstctlcite{IEEEauthorcontrol}

\maketitle
\thispagestyle{pacepreprint}

\begin{abstract}
As multimodal agents accumulate experience, their long-term memories contain many similar observations of recurring scenes, actions, and entities. Answer quality depends on the evidence retrieved from these memories into the model's context. However, query-relevant memories may contain overlapping information, and a fixed retrieval depth may return too little or too much evidence for a given question. To address these challenges, we propose PACE-Mem, Prefix-Aware Complementary Evidence, which constructs a complementary evidence sequence and determines its length from the same selection trajectory. For structured memories, query-guided pooling combines records within each clip to obtain relevance scores and representations. Using these relevance scores and representations, maximal marginal relevance (MMR) selection balances query relevance against overlap with previously selected evidence. From the resulting score trajectory, a prefix-length selection algorithm determines retrieval depth by comparing length-normalized adjacent declines, accounting for both score changes and the amount of evidence accumulated. PACE-Mem is a training-free, plug-in retrieval method for clip-based agent memory systems, leaving memory construction and answer generation unchanged. Across three long-video benchmarks and two agent frameworks, PACE-Mem improves question-answering accuracy by an average of 1.6\% over Adaptive-$k$, the strongest baseline at comparable mean retrieval depths. Further analysis reveals that jointly adding and removing evidence is associated with larger net accuracy gains than either change alone.
\end{abstract}


\begin{figure}[!t]
  \centering
  \includegraphics[width=\linewidth]{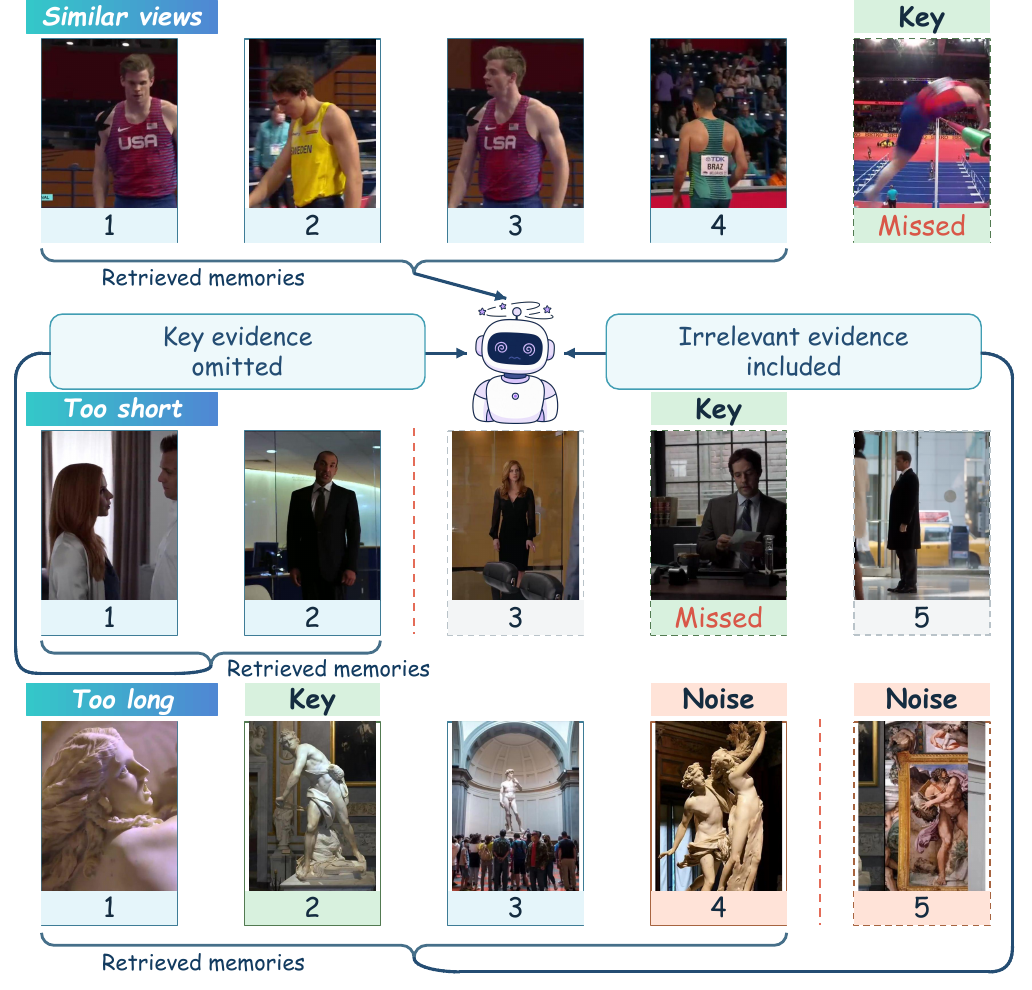}
  \caption{Evidence selection in long-video memory, illustrated with VideoMME-Long frames. Top: retrieved views repeat nearby activity around a pole-vault attempt while omitting the view that reveals contact with the bar. Middle: a short prefix misses the check handoff. Bottom: a long prefix includes unrelated artwork after the key view. PACE-Mem chooses prefix length from the scores obtained while selecting complementary clips.}
  \label{fig:teaser}
\end{figure}

\section{Introduction}
\label{sec:introduction}

\IEEEPARstart{M}{ultimodal} agents use long-term memory to answer questions about visual events and speech encountered in past experiences~\cite{long2026m3agent,zhang2026silvr,ren2026icmbench}. As interactions accumulate, this memory preserves information beyond the agent's current observation. Retrieval brings relevant experiences into the model's context~\cite{park2023generativeagents,zhong2024memorybank}, making the evidence available at answering time depend on which parts of the stored history are returned.

Agent memory retrieval operates over an accumulated stream of experience, where relevant records can be strongly correlated. This overlap within a shared history distinguishes memory retrieval from search over heterogeneous document collections~\cite{hu2026xmemory}. In multimodal memory, adjacent video intervals often depict the same scene, people, or ongoing action. An event may appear in both visual descriptions and speech transcripts, and structured memory may associate multiple episodic and semantic records with one interval. Ranking records independently by relevance can therefore return repeated content while overlooking a decisive detail. In the top example of Fig.~\ref{fig:teaser}, relevance-only retrieval returns similar pole-vault views but misses the view revealing contact with the bar. Evidence needs also vary by question: a short context may omit necessary facts, while a long one can make relevant information difficult to use~\cite{liu2024lostmiddle}. The middle and bottom examples of Fig.~\ref{fig:teaser} illustrate a missed check handoff and the inclusion of visually salient but unrelated artwork, respectively.

These examples show why retrieval must determine both which records to return and how many. A longer context can preserve existing evidence while adding new clips, whereas changing its composition can introduce a missing fact and remove competing information without increasing its size.

Existing work addresses agent memory through memory-system design and retrieval-time evidence selection. xMemory builds a revisable hierarchy of memory components and groups, then retrieves through this hierarchy with uncertainty-guided evidence expansion~\cite{hu2026xmemory}. A-MEM organizes memories as interconnected notes and updates their links and contextual attributes as new experiences arrive~\cite{xu2025amem}. A complementary direction develops retrieval methods that control which evidence enters the answer context. Adaptive-RAG uses a question-complexity classifier to choose among no retrieval, single-step retrieval, and iterative retrieval~\cite{jeong2024adaptiverag}. Adaptive-$k$ locates the largest gap in ranked query--document relevance scores and extends the resulting cutoff with a fixed buffer~\cite{taguchi2025adaptivek}. Redundancy-aware selection has also been combined with quantity control, using relative declines between consecutive maximal marginal relevance (MMR) scores or gaps between utilities that account for redundancy~\cite{lu2014microblog,guo2026adagate}.

Improved memory organization alone does not ensure complementary evidence or a suitable retrieval depth. Even structured memories can retain overlapping observations of the same scene from similar viewpoints. Fixed-depth retrieval cannot adapt to varying evidence needs, while question-based routing and initial relevance-gap cutoffs do not track overlap with subsequently selected evidence. Redundancy-aware selection accounts for this overlap, but adjacent score declines alone do not explicitly reflect the amount of evidence already selected. The remaining question is how to compare candidate stopping points that correspond to different prefix lengths.

To address this problem, we propose PACE-Mem, \emph{Prefix-Aware Complementary Evidence}, which determines evidence composition and retrieval depth from a shared selection trajectory. For structured memories, query-guided pooling combines related records within each clip to obtain relevance scores and representations for ranking and redundancy assessment. Flat memories use their existing clip representations. MMR~\cite{carbonell1998mmr} then constructs an evidence sequence by balancing query relevance against overlap with earlier selections. From the recorded scores, our prefix-length selection algorithm chooses the initial portion of this sequence to return. It compares adjacent score declines after length normalization, accounting for both local score changes and the amount of evidence accumulated. PACE-Mem is training-free and can be plugged into these memory interfaces without changing memory construction or answer generation. Evidence selection requires no additional foundation-model calls.

We evaluate PACE-Mem with M3-Agent and a retrieval-based SiLVR configuration on M3Bench Robot, M3Bench Web, and VideoMME-Long. Average accuracy rises from 44.2\% with Original M3-Agent to 52.2\% with PACE-Mem, an 8.0\% gain. Across the two frameworks and three benchmarks, PACE-Mem improves accuracy by an average of 1.6\% over Adaptive-$k$, the strongest baseline at comparable mean retrieval depths.

Further evidence-set analysis shows that similar retrieval depths can correspond to different evidence compositions. In SiLVR's single-retrieval setting, questions for which PACE-Mem both adds and removes clips relative to Fixed-7 exhibit larger within-group net accuracy gains than those involving only additions or only removals. This finding highlights the importance of evidence composition: retrieval must determine not only how much evidence to return, but also which evidence to retain, introduce, and remove.

Our contributions are as follows:
\begin{itemize}
  \item We formulate memory retrieval as the joint selection of evidence composition and retrieval depth, where the returned amount depends on the complementary evidence accumulated for each question.
  \item We propose PACE-Mem, a training-free, plug-in method that combines complementary sequence construction with length-normalized prefix-length selection. For structured memories, query-guided pooling supplies relevance scores and representations for both candidate ranking and redundancy assessment.
  \item We evaluate PACE-Mem in two agent frameworks across three long-video benchmarks, covering structured and flat memory interfaces. Paired evidence-set analysis relates membership changes to answer corrections and regressions, complementing aggregate accuracy and retrieval-depth comparisons.
\end{itemize}

\section{Related Work}
\label{sec:related}

\begin{figure*}[!t]
  \centering
  \includegraphics[width=0.99\textwidth]{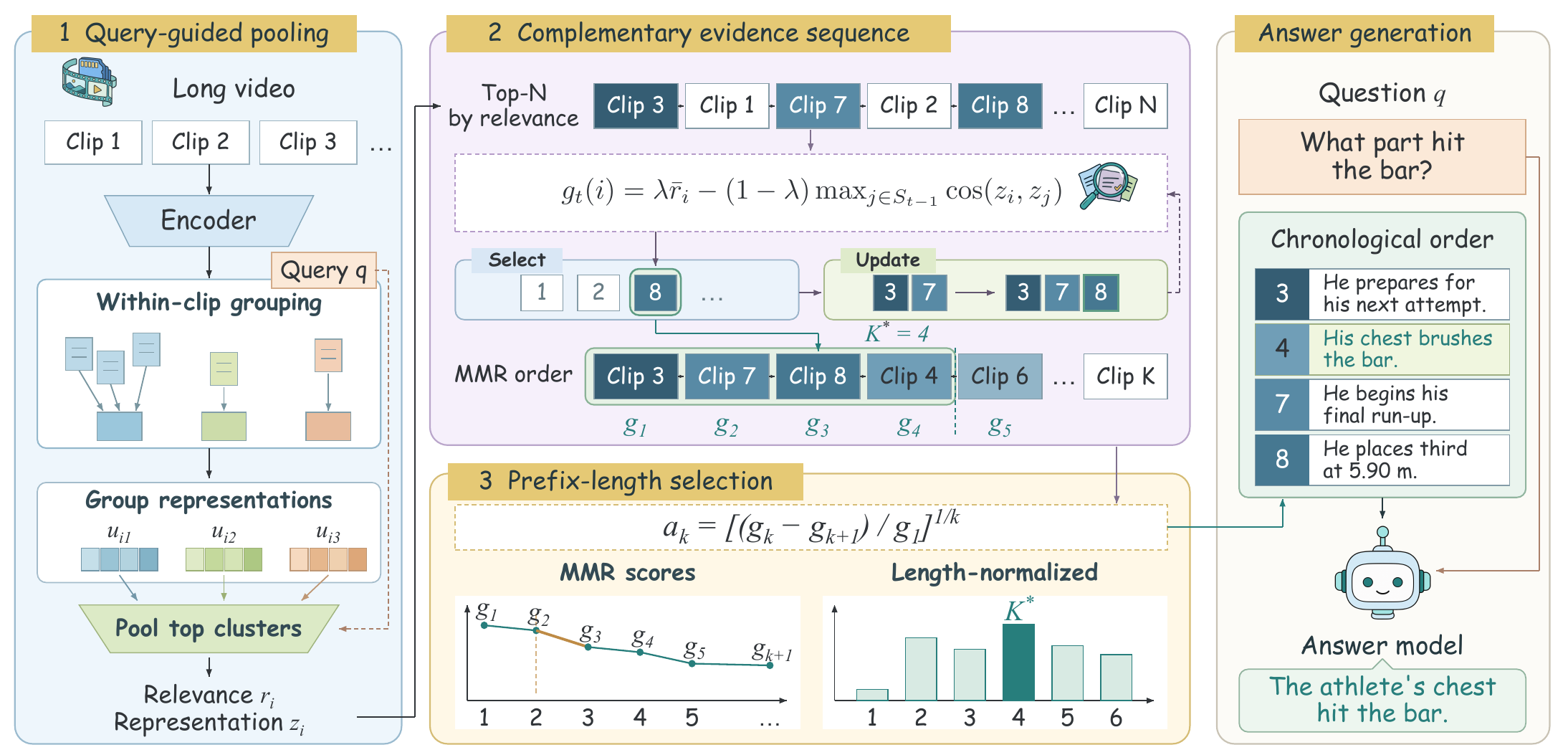}
  \caption{Retrieval with PACE-Mem. The host encodes video into textual memory. For structured memories, similar records within each clip are grouped before query-guided pooling produces relevance scores and evidence representations. Consistent blue shades track clip identities across the relevance ranking and MMR sequence. The largest raw decline occurs at $k=2$, while the length-normalized criterion selects $K^*=4$. The answer model receives the selected prefix in chronological order. The question, answer, and memory excerpts are adapted from VideoMME-Long question 769-2; clip IDs and scores are schematic.}
  \label{fig:method}
\end{figure*}

\subsection{Agent Memory}

Agent memory systems organize, update, and retrieve accumulated experience to support later interactions~\cite{gutierrez2024hipporag}. Generative Agents and MemoryBank retain observations or conversational information, while Reflexion stores feedback to guide subsequent attempts~\cite{park2023generativeagents,zhong2024memorybank,shinn2023reflexion}. MemGPT and MemoryOS manage growing histories across memory tiers~\cite{packer2023memgpt,kang2025memoryos}. To update content, Mem0 reconciles extracted conversational facts with existing memories, while A-MEM links new notes to prior records and revises their contextual attributes~\cite{chhikara2025mem0,xu2025amem}. xMemory addresses correlated records through hierarchical organization and uncertainty-guided expansion~\cite{hu2026xmemory}. For multimodal experience, M3-Agent links episodic and semantic records around entities and retrieves them during iterative reasoning~\cite{long2026m3agent}. SiLVR~\cite{zhang2026silvr} provides video descriptions that form the flat evidence store in our retrieval-based configuration. Memory organization alone does not determine which overlapping records to return for a particular question. PACE-Mem selects complementary evidence and its quantity over these structured and flat interfaces without changing memory construction.

\resultsubsection{Long-Video Reasoning}

Long-video reasoning requires identifying and integrating relevant evidence across time under limited input capacity~\cite{qian2024locans}. M3Bench, Video-MME, and related benchmarks evaluate reasoning over video content and long-term multimodal experience~\cite{long2026m3agent,fu2025videomme,ren2026icmbench}. Visual--language models provide alignment and video interaction capabilities~\cite{radford2021clip,li2023blip2,zhang2023videollama}, while MovieChat, Flash-VStream, and LongVU extend temporal coverage through memory or token compression~\cite{song2024moviechat,zhang2025flashvstream,shen2025longvu}. Language-based approaches include concept-based video captioning~\cite{wang2022crossmodalgraph,yang2023conceptaware}, and LLoVi reasons over descriptions of local intervals~\cite{zhang2024llovi}. For evidence selection, Video Pr\'ecis favors diverse content under a summary-length constraint, and multi-view summarization combines joint embeddings with sparse optimization~\cite{shroff2010precis,panda2017multiview}. For question answering and localization, spatiotemporal or event graphs model relevant relations~\cite{cheng2024krst,bai2024eventgraph,wang2024hsshg}, while language hierarchies and caption-assisted retrieval identify target moments~\cite{ding2022languagehierarchy,xie2025captionassisted}. R-VLM learns to retrieve a prescribed number of video chunks~\cite{xu2026rvlm}. UniVid revises the evidence set using evaluator feedback and MMR within Pyramid Reflection to expand or prune frames~\cite{luo2025univid}. A prescribed retrieval size cannot adapt to varying evidence needs, while feedback-driven revision requires evaluating intermediate selections. PACE-Mem selects complementary clips and their quantity from stored memories using a shared score trajectory, without additional foundation-model calls.

\resultsubsection{Adaptive Retrieval}

Adaptive retrieval controls evidence access in retrieval-augmented generation~\cite{lewis2020rag,karpukhin2020dpr,izacard2022contriever,khattab2020colbert,izacard2021fid} using query characteristics, model feedback, or retrieval scores. Query and corpus features predict document counts~\cite{kratzwald2018adaptive}, and Adaptive-RAG routes questions by complexity~\cite{jeong2024adaptiverag}. FLARE, Self-RAG, and CRAG use generation confidence, learned self-assessment, or evidence evaluation to guide retrieval~\cite{jiang2023flare,asai2024selfrag,yan2024crag}. Score-based control includes Adaptive-$k$'s relevance-gap cutoff with a buffer and EviRerank's threshold relative to the top-scoring block~\cite{taguchi2025adaptivek,li2026evirerank}. A cutoff derived from the initial relevance ranking does not reflect how a candidate's novelty changes as evidence is selected. This motivates coupling quantity control to a redundancy-aware selection sequence. MMR penalizes overlap with earlier selections~\cite{carbonell1998mmr}, and DF-RAG adjusts its geometric variant's diversity weight through model-based planning and evaluation~\cite{khan2026dfrag}. Extending redundancy-aware selection to quantity control, MMR-based microblog retrieval uses relative score declines between consecutive results as a stopping criterion~\cite{lu2014microblog}. AdaGReS selects evidence with positive redundancy-aware gains under a token budget~\cite{peng2025adagres}, whereas AdaGATE sets capacity from the largest gap in utilities incorporating redundancy and information-gap coverage~\cite{guo2026adagate}. Relative score declines capture local changes but do not explicitly include the length of the preceding evidence prefix. PACE-Mem incorporates this length through a prefix-dependent power transformation of declines normalized by the initial score. This adjusts the comparison of candidate stopping positions according to the amount of evidence already selected.

\section{Prefix-Aware Complementary Evidence}
\label{sec:method}

\subsection{Overview}
\label{sec:overview}

PACE-Mem couples complementary evidence selection with adaptive retrieval depth through three connected stages (Fig.~\ref{fig:method}). For structured memories, query-guided pooling groups related records within each clip and aggregates query-relevant groups into a relevance score and representation. MMR uses these outputs to construct an evidence sequence, balancing query relevance against overlap with previously selected clips. Prefix-length selection uses the same MMR score trajectory to determine how many clips to return by comparing length-normalized score declines. Records from the selected prefix are supplied chronologically to the answer model. Algorithm~\ref{alg:retrieval} summarizes the procedure.

\resultsubsection{Problem Setup}
\label{sec:problem_setup}

Memory retrieval determines which evidence to return and how much to include for a question. These decisions are coupled because the usefulness of additional evidence depends on what is already selected. Given a query $q$ and accumulated memory $\mathcal M$, the host provides $T$ temporal evidence units $\mathcal X=\{x_i\}_{i=1}^{T}$, each associated with stored records and a temporal identifier. In our video setting, an evidence unit is a clip, which can contain multiple memory records. For each unit $x_i$, we compute query relevance $r_i$ and use an $\ell_2$-normalized representation $z_i$ to assess overlap. Retrieval selects an ordered sequence of unit indices $(i_1,\ldots,i_K)$, with $S_K=\{i_1,\ldots,i_K\}$ denoting its index set and $K$ its returned length. The corresponding records are supplied chronologically to the answer model $F$, which predicts $\hat y$.

\resultsubsection{Query-Guided Pooling}
\label{sec:candidate-construction}

PACE-Mem obtains clip-level relevance scores and representations through query-guided pooling for structured memories or directly from flat memories. In M3-Agent, multiple episodic and semantic records, stored as memory nodes, describe one interval~\cite{long2026m3agent}. Pooling combines their most query-relevant groups into a clip-level score and representation, carrying information from related records forward to the subsequent selection stage. In our retrieval-based SiLVR configuration~\cite{zhang2026silvr}, each 30-s interval is represented by one document containing a visual description and automatic speech recognition (ASR) transcripts. Its stored $\ell_2$-normalized embedding supplies $z_i$, and cosine similarity to the query supplies $r_i$.

Within-clip grouping clusters similar query-relevant records. For clip $i$ with $n_i$ records, $v_{ij}$, $s_{ij}$, and $c_{ij}$ denote record $j$'s normalized embedding, query score, and memory type. The score $s_{ij}$ follows the host's node retrieval: the maximum cosine similarity over query variants and the record's stored embeddings. We cluster up to $H$ highest-scoring records in descending score order. Each joins the most similar existing cluster if its cosine similarity to the centroid exceeds $\tau$, or starts a new cluster otherwise. For cluster index set $G_{i\ell}$, we use weights $w_{ij}=\max(s_{ij},\eta)$ with a positive floor $\eta>0$ to form the centroid
\begin{equation}
 \mu_{i\ell}=\operatorname{norm}\!\left(\sum_{j\in G_{i\ell}}w_{ij}v_{ij}\right),
 \label{eq:group-summary}
\end{equation}
where $\operatorname{norm}$ denotes $\ell_2$ normalization. The centroid is updated after each record joins the cluster.

Query-guided aggregation combines the leading clusters into a clip-level relevance score and representation. We rank the $m_i$ clusters by their strongest query scores $u_{i\ell}=\max_{j\in G_{i\ell}}s_{ij}$ and pool the leading $L_i=\min(L,m_i)$ clusters:
\begin{IEEEeqnarray}{rCl}
 r_i &=&\sum_{\ell=1}^{L_i}\alpha_{\ell}u_{i\ell}
      +\beta_{\rm type}b_i^{\rm type}
      +\beta_{\rm cov}\min(m_i-1,C),\nonumber\\
 z_i &=&\operatorname{norm}\!\left(
 \sum_{\ell=1}^{L_i}\alpha_\ell\max(u_{i\ell},\eta)\mu_{i\ell}
 \right).
\IEEEeqnarraynumspace
\label{eq:cliprepr}
\end{IEEEeqnarray}
Here $L$ caps the pooled clusters, and decreasing rank weights $\alpha_\ell$ reduce weaker matches' contributions. The indicator $b_i^{\rm type}\in\{0,1\}$ is one when records in the leading $\min(P,m_i)$ clusters contain both episodic and semantic types. Thus, $P$ controls the type check, while $C$ caps the additional clusters rewarded. The coefficients $\beta_{\rm type}$ and $\beta_{\rm cov}$ weight these bonuses. With cluster membership fixed, lower-scoring records cannot increase the maximum score, but their embeddings contribute to the centroid used for overlap assessment.

Candidate admission retains the top-scoring clips and normalizes their relevance for complementary selection. We recompute pooling from stored embeddings, retaining node order for score ties and the earliest cluster for similarity ties. We keep at most $N$ clips, with candidate index set $\mathcal C_q=\operatorname{TopN}_{i\in\{1,\ldots,T\}}r_i$. An empty pool returns no evidence. Otherwise, we rescale relevance as
\begin{equation}
 \bar r_i=\frac{r_i-r_{\min}}{\max(r_{\max}-r_{\min},10^{-12})},
 \label{eq:relevance-scale}
\end{equation}
where $r_{\min}$ and $r_{\max}$ are the pool's minimum and maximum scores. This preserves ranking and places relevance in $[0,1]$ for combination with cosine similarity in $[-1,1]$.

\subsection{Complementary Evidence Sequence}
\label{sec:mmr}

Complementary sequence construction produces an evidence ordering and a score trajectory for subsequent depth selection. Selection balances query relevance against overlap with earlier selections. Starting from $S_0=\varnothing$, step $t$ selects the index $i_t=\arg\max_{i\in\mathcal C_q\setminus S_{t-1}}g_t(i)$ using
\begin{equation}
 g_t(i)=
 \begin{cases}
   \bar r_i, & t=1,\\[4pt]
   \displaystyle\lambda\bar r_i-(1-\lambda)
   \max_{j\in S_{t-1}}\cos(z_i,z_j), & t>1.
 \end{cases}
 \label{eq:mmr}
\end{equation}
The first clip is chosen by relevance alone. Subsequent steps use classical MMR~\cite{carbonell1998mmr}, where $\lambda\in[0,1]$ balances relevance against maximum similarity to an earlier selection. We update $S_t=S_{t-1}\cup\{i_t\}$ and retain the selected score $g_t=g_t(i_t)$. Each $g_t$ is thus conditioned on the preceding selections.

The maximum-similarity penalty accounts for overlap with any previously selected clip. Averaging similarities over the selected set could dilute a strong match to one clip when the other selected clips are unrelated, making repeated content less visible in the resulting penalty. Within-clip grouping summarizes related records within each interval, while this cross-clip penalty discourages selecting similar evidence from different intervals.

From the second selection onward, the scores are non-increasing. Relevance scores, representations, and $\lambda$ remain fixed during sequence construction. As $S_{t-1}$ expands to $S_t$, each remaining candidate's maximum similarity can only increase or stay unchanged. Since $1-\lambda\geq0$, its score therefore satisfies $g_{t+1}(i)\leq g_t(i)$ for $t\geq2$. The next selected candidate was also available at step $t$, giving $g_{t+1}=g_{t+1}(i_{t+1})\leq g_t(i_{t+1})\leq g_t$. The first transition follows separately when min-max scaling gives $g_1=1$. With $\bar r_i\leq1$ and $\cos(z_i,z_j)\geq-1$, we have $g_2\leq\lambda+(1-\lambda)=1=g_1$.

\subsection{Prefix-Length Selection}
\label{sec:depth}

Prefix-length selection uses a length-normalized score decline to choose a boundary on the complementary sequence. Adaptive-$k$ compares gaps between independently ranked relevance scores and adds a fixed buffer~\cite{taguchi2025adaptivek}. Evaluating boundaries on the MMR trajectory instead couples depth selection to evidence composition: an earlier near-duplicate can change a candidate's score, position, and adjacent declines. For a positive, non-increasing trajectory, the adjacent-score ratio $p_k=g_{k+1}/g_k$ measures how much retrieval value the next clip retains relative to the current selection. A ratio near one supports continuation because the next clip remains similarly valuable under the relevance--overlap criterion, whereas a small ratio indicates a sharp decline and supports truncation. This relative measure is invariant to a common positive rescaling of the score trajectory and lies in $(0,1]$, motivating its use as a local continuation probability. Stopping after clip $k$ requires $k-1$ continuations followed by a stop, yielding the probability $\Pr(K=k)$ and its length-normalized score $a_k$:
\begin{IEEEeqnarray}{rCl}
 \Pr(K=k)
 &=&\left(\prod_{i=1}^{k-1}p_i\right)(1-p_k)
 =\frac{g_k-g_{k+1}}{g_1},\nonumber\\
 a_k &=&\Pr(K=k)^{1/k}
 =\left(\frac{g_k-g_{k+1}}{g_1}\right)^{1/k}.
\IEEEeqnarraynumspace
\label{eq:root}
\end{IEEEeqnarray}

\begin{figure}[!t]
\refstepcounter{algorithm}\label{alg:retrieval}
\hrule\smallskip
\noindent\textbf{Algorithm~\thealgorithm} PACE-Mem evidence selection
\par\smallskip\hrule\smallskip
\small
\begin{algorithmic}[1]
\REQUIRE Query $q$, memory $\mathcal M$, $N$, $K_{\min}$, $K_{\max}$, $\lambda$
\ENSURE Selected memory records in chronological order
\STATE Form clip candidates $\mathcal X$ (Section~\ref{sec:candidate-construction})
\STATE $\mathcal C_q\leftarrow\operatorname{TopN}_{i\in\{1,\ldots,T\}}r_i$
\STATE \textbf{if} $\mathcal C_q=\varnothing$ \textbf{then return} $\varnothing$
\STATE Rescale relevance as in Section~\ref{sec:candidate-construction}
\STATE $K_{\rm ret}\leftarrow\min(K_{\max},|\mathcal C_q|)$
\STATE $K_{\rm obs}\leftarrow\min(K_{\rm ret},|\mathcal C_q|-1)$\quad $S_0\leftarrow\varnothing$
\FOR{$t=1,\ldots,K_{\rm obs}+1$}
  \STATE $i_t\leftarrow\arg\max_{i\in\mathcal C_q\setminus S_{t-1}}g_t(i)$ using \eqref{eq:mmr}
  \STATE $g_t\leftarrow g_t(i_t)$\quad $S_t\leftarrow S_{t-1}\cup\{i_t\}$
\ENDFOR
\STATE $\mathcal K_+\leftarrow\{k\in\{1,\ldots,K_{\rm obs}\}:g_k>g_{k+1}\}$
\STATE $\widetilde K\leftarrow K_{\rm ret}$
\IF{$g_1>0$ and $\mathcal K_+\ne\varnothing$}
  \STATE $\widetilde K\leftarrow\arg\max_{k\in\mathcal K_+}[(g_k-g_{k+1})/g_1]^{1/k}$
\ENDIF
\STATE $K^*\leftarrow\min(K_{\rm ret},\max(K_{\min},\widetilde K))$
\RETURN First $K^*$ MMR-selected clips in temporal order
\end{algorithmic}
\smallskip\hrule
\end{figure}

Length normalization accounts for the different numbers of factors underlying candidate stopping positions. A prefix of length $k$ involves $k-1$ continuation factors and one stopping factor. Multiplying additional continuation factors in $(0,1]$ reduces or preserves their cumulative product, which can disadvantage longer prefixes when stopping factors are comparable. The full boundary probability also depends on the stopping factor, so its ranking reflects both effects. Taking the $k$th root converts the product of $k$ factors into their geometric mean: $\log a_k=k^{-1}[\sum_{i=1}^{k-1}\log p_i+\log(1-p_k)]$ for positive boundary probabilities. This compares the average log score per decision across candidate lengths, a motivation related to length normalization in sequence generation~\cite{murray2018lengthbias}. Without the root, maximizing $\Pr(K=k)$ simply maximizes $g_k-g_{k+1}$ because $g_1>0$ is shared across boundaries; on a relevance-ranked trajectory, this recovers Adaptive-$k$'s largest-gap criterion before its buffer is added. For an equal normalized decline in $(0,1)$, it favors a longer prefix: a decline of $0.16$ gives $a_2=0.40$ and $a_4\approx0.63$ (Fig.~\ref{fig:method}). More generally, writing $\delta_k=(g_k-g_{k+1})/g_1$ for an eligible boundary, comparison of two positive normalized declines gives
\begin{equation}
 a_\ell>a_k
 \quad\Longleftrightarrow\quad
 \frac{\log\delta_\ell}{\ell}>\frac{\log\delta_k}{k}
 \quad\Longleftrightarrow\quad
 \delta_\ell>\delta_k^{\ell/k}.
 \label{eq:boundary-comparison}
\end{equation}
For $\ell>k$ and $0<\delta_k<1$, the threshold is smaller than $\delta_k$, allowing a later boundary to win with a smaller decline. Unlike a fixed buffer, this preference depends on both the decline and the relative prefix lengths.

Final selection applies the boundary score within the host's depth bounds and available candidates. The decline-based score remains well-defined whenever $g_1>0$ and $g_k>g_{k+1}$, including when later scores are nonpositive. We therefore compute it directly from the normalized decline, extending the selection rule beyond the positive trajectories used in the probability derivation. Each observable boundary requires a successor, so the host's bounds $K_{\min}$ and $K_{\max}$ give
\begin{IEEEeqnarray}{rCl}
 K_{\rm ret} &=& \min(K_{\max},|\mathcal C_q|),\nonumber\\
 K_{\rm obs} &=& \min(K_{\rm ret},|\mathcal C_q|-1).
 \label{eq:depth-limits}
\end{IEEEeqnarray} We compare boundaries from the first position onward, defining $\mathcal K_+=\{k\in\{1,\ldots,K_{\rm obs}\}:g_k>g_{k+1}\}$. If $g_1>0$ and $\mathcal K_+\ne\varnothing$, we select $\widetilde K=\arg\max_{k\in\mathcal K_+}a_k$; otherwise, $\widetilde K=K_{\rm ret}$. This fallback handles equal relevance scores, which yield $g_1=0$ under min-max scaling, and trajectories without a strictly decreasing boundary. The lower bound preserves the host's minimum context size, while the upper bound limits the returned evidence to its budget and available candidates. The final length is
\begin{equation}
 K^*=\min\!\left(K_{\rm ret},\max(K_{\min},\widetilde K)\right).
 \label{eq:bounded-depth}
\end{equation} If fewer than $K_{\min}$ candidates exist, all are returned. The bounded trajectory is constructed before prefix selection, including a successor when needed to evaluate $K_{\max}$. Although evaluating a boundary requires inspecting the following clip, that successor supplies only its comparison score and is excluded from the returned prefix at that boundary. The answer is $\hat y=F(q,\operatorname{order}_{\rm time}(S_{K^*}))$, where temporal sorting changes only presentation order, not the selected members or boundary scores.


\section{Experiments}
\label{sec:experiments}

\subsection{Experimental Setup}
\label{sec:experimental_setup}

\begin{table*}[!tp]
\caption{Question-answering accuracy and mean retrieval depth in two agent frameworks. Clips reports the mean evidence count per nonempty ordinary retrieval call. Results are compared within each host block, with SiLVR Web evaluated on a fixed subset of 3,171 questions}
\label{tab:main}
\centering
\definecolor{gaingreen}{HTML}{277E65}
\definecolor{lossred}{HTML}{B5513E}
\definecolor{deltagray}{HTML}{647178}
\newcommand{\gain}[1]{\textsubscript{\textcolor{gaingreen}{\scriptsize$\uparrow$#1}}}
\newcommand{\loss}[1]{\textsubscript{\textcolor{lossred}{\scriptsize$\downarrow$#1}}}
\newcommand{\same}[1]{\textsubscript{\textcolor{deltagray}{\scriptsize$\rightarrow$#1}}}
\setlength{\tabcolsep}{4pt}
\renewcommand{\arraystretch}{1.14}
\footnotesize
\begin{tabularx}{\textwidth}{@{}>{\columncolor{white}[0pt][\tabcolsep]}l*{6}{Y}>{\columncolor{white}[\tabcolsep][0pt]}Y@{}}
\toprule
& \multicolumn{2}{c}{\textbf{M3Bench Robot}} &
\multicolumn{2}{c}{\textbf{M3Bench Web}} &
\multicolumn{2}{c}{\textbf{VideoMME-Long}} & \\
\cmidrule(lr){2-3}\cmidrule(lr){4-5}\cmidrule(lr){6-7}
\multirow{-2}{*}{\textbf{Method}} & Acc. (\%) & Clips & Acc. (\%) & Clips & Acc. (\%) & Clips & \multirow{-2}{*}{\textbf{Avg. Acc.}}\\
\midrule
\multicolumn{8}{c}{\textbf{M3-Agent}\quad\textit{Answer model: M3-Agent-Control (Qwen3-32B)}}\\
\midrule
Original & 30.7 & 1.7 & 48.9 & 1.8 & 53.0 & 1.8 & 44.2\\
Fixed-4 & 38.4 & 3.9 & 56.4 & 4.0 & 56.9 & 4.0 & 50.6\\
Fixed-5 & 38.9\gain{0.5} & 4.8 & 56.7\gain{0.3} & 5.0 & 57.6\gain{0.7} & 5.0 & 51.1\gain{0.5}\\
Adaptive-$k$ & 38.8\gain{0.4} & 4.0 & 55.3\loss{1.1} & 4.1 & 57.9\gain{1.0} & 4.1 & 50.7\gain{0.1}\\
Adaptive-RAG & 38.3\loss{0.1} & 4.0 & 55.1\loss{1.3} & 4.2 & 56.8\loss{0.1} & 4.1 & 50.1\loss{0.5}\\
\rowcolor{rowblue}\textbf{PACE-Mem (ours)} & \textbf{40.3}\gain{1.9} & 3.9 & \textbf{57.6}\gain{1.2} & 4.1 & \textbf{58.7}\gain{1.8} & 4.3 & \textbf{52.2}\gain{1.6}\\
\midrule
\multicolumn{8}{c}{\textbf{SiLVR}\quad\textit{Answer model: Qwen3-VL-8B-Instruct}}\\
\midrule
Fixed-7 & 40.8 & 7.0 & 55.9 & 7.0 & 56.9 & 7.0 & 51.2\\
Fixed-8 & 41.3\gain{0.5} & 8.0 & \textbf{57.6}\gain{1.7} & 8.0 & 58.0\gain{1.1} & 8.0 & 52.3\gain{1.1}\\
Adaptive-$k$ & 41.4\gain{0.6} & 6.8 & 56.1\gain{0.2} & 6.8 & 56.2\loss{0.7} & 6.9 & 51.2\same{0.0}\\
Adaptive-RAG & 41.0\gain{0.2} & 7.0 & 56.3\gain{0.4} & 7.2 & 57.2\gain{0.3} & 7.1 & 51.5\gain{0.3}\\
\rowcolor{rowblue}\textbf{PACE-Mem (ours)} & \textbf{42.4}\gain{1.6} & 7.1 & 57.2\gain{1.3} & 7.0 & \textbf{59.0}\gain{2.1} & 7.2 & \textbf{52.9}\gain{1.7}\\
\bottomrule
\end{tabularx}
\par\smallskip
\begin{minipage}{\textwidth}\scriptsize
Accuracy changes (percentage points) are relative to Fixed-4 for M3-Agent and Fixed-7 for SiLVR ($\rightarrow$: unchanged). Bold marks the best accuracy per framework.
\end{minipage}
\end{table*}

\textbf{Datasets.} We evaluate PACE-Mem on M3Bench Robot, M3Bench Web~\cite{long2026m3agent}, and the long-video subset of Video-MME, referred to as VideoMME-Long~\cite{fu2025videomme}. Robot comprises 1,276 open-ended questions from 100 embodied indoor videos, while Web comprises 3,214 questions from 920 YouTube videos. VideoMME-Long contains 900 multiple-choice questions from 300 videos. M3-Agent uses the released questions and memories, whereas SiLVR memories are reconstructed from the source videos. Since 14 Web videos were unavailable, SiLVR is evaluated on a fixed subset of 906 videos and 3,171 questions. Within each host configuration, all methods use the same questions.

\textbf{Baselines.} We compare PACE-Mem with fixed-depth retrieval and two adaptive context controllers within each host framework, alongside Original M3-Agent. M3-Agent retrieves during ordinary memory searches under its existing control policy, whereas the SiLVR configuration retrieves once from flat caption and ASR memory. Original M3-Agent retains its released relevance threshold and chain-of-retrieval behavior. The other M3-Agent variants remove this threshold and backfill candidates to share a common candidate-admission policy. Fixed-$k$ returns the highest-ranked clips by dense relevance. Fixed-4 in M3-Agent and Fixed-7 in SiLVR provide comparisons at similar mean retrieval depths, while Fixed-5 and Fixed-8 provide additional comparisons with larger returned contexts. Adaptive-$k$ selects a relevance-gap boundary and appends a buffer~\cite{taguchi2025adaptivek}. Adaptive-RAG uses its released T5-large complexity classifier without retraining~\cite{jeong2024adaptiverag}; we map its three retrieval-strategy classes to context depths while retaining each host's reasoning loop. The buffers and class mappings approximate PACE-Mem's mean returned depth within each host and remain fixed across benchmarks.

\textbf{Metrics.} We evaluate answer quality using accuracy and returned evidence quantity using Clips. Accuracy is determined by exact option matching on VideoMME-Long and GPT-4o correctness judgments on M3Bench~\cite{openai2024gpt4o}. All methods use the same judging prompt, containing the question, reference answer, and generated answer; method identities and retrieval budgets are omitted. Avg. Acc. is the unweighted mean accuracy across the three benchmarks, rounded to one decimal place. For the overall comparison, we select the single baseline with the highest mean accuracy across both frameworks and all three benchmarks among those at comparable mean retrieval depths. Overall gains are averaged across these six evaluations before rounding. Clips is the mean number of evidence clips returned per nonempty ordinary retrieval call, excluding M3-Agent's entity lookups and other searches. It measures evidence quantity at the retrieval-call level, where record lengths and the number of calls per question can vary. The realized depth can fall below the requested depth when candidates are exhausted.

\begingroup\clubpenalty=10000\looseness=2
\textbf{Implementation details.} We retain each host's model configuration and use fixed retrieval parameters across benchmarks. M3-Agent uses the released M3-Agent-Control (DAPO-trained Qwen3-32B) with its official decoding settings and \texttt{text-embedding-3-large} for query encoding~\cite{long2026m3agent}. To construct SiLVR memories, we divide source videos into nonoverlapping 30-s intervals. Qwen3-VL-8B-Instruct~\cite{bai2025qwen3vl} generates descriptions from each interval and its timestamped transcript, and Qwen3-Embedding-4B~\cite{zhang2025qwen3embedding} encodes the caption and ASR records. SiLVR uses Qwen3-VL-8B-Instruct with the official task-specific answer prompts~\cite{zhang2026silvr} and greedy decoding. Within each host, methods share the evidence store, clip boundaries, query representations, dense search, answer model, prompts, and decoding settings. The answer model receives the question together with the selected records ordered chronologically. Both hosts use $N=200$, $\lambda=0.85$, and $K_{\min}=2$. We set $K_{\max}=5$ for M3-Agent, which can request further evidence in subsequent calls, and $K_{\max}=8$ for SiLVR's single returned context. Query-guided pooling in M3-Agent uses $H=8$, $L=4$, $P=3$, $\tau=0.85$, $\eta=10^{-3}$, $\boldsymbol\alpha=(1,0.35,0.15,0.05)$, $\beta_{\rm type}=0.05$, $\beta_{\rm cov}=0.03$, and $C=3$. Original M3-Agent uses its released relevance threshold of 0.5. Adaptive-$k$ appends two clips in M3-Agent and five in SiLVR. Adaptive-RAG's classes for no retrieval, single retrieval, and iterative retrieval map to depths 2/4/5 in M3-Agent and 2/7/8 in SiLVR. Its classifier receives only the question, whereas multiple-choice retrieval queries include all answer options.\par\endgroup

\begin{table}[!t]
\caption{Component ablations in SiLVR on VideoMME-Long. Checks indicate enabled components. Removing MMR uses relevance order, and removing depth selection fixes $K=7$. Clips is the mean returned depth}
\label{tab:components}
\centering
\setlength{\tabcolsep}{3pt}
\renewcommand{\arraystretch}{1.08}
\begin{tabularx}{\columnwidth}{@{}>{\columncolor{white}[0pt][\tabcolsep]}lc>{\centering\arraybackslash}Xc>{\columncolor{white}[\tabcolsep][0pt]}c@{}}
\toprule
\textbf{Variant} & \textbf{MMR} & \textbf{Depth selection} & \textbf{Acc. (\%)} & \textbf{Clips}\\
\midrule
\rowcolor{rowblue} PACE-Mem & $\checkmark$ & $\checkmark$ & \textbf{59.0} & 7.2\\
w/o MMR & -- & $\checkmark$ & 58.1 & 7.4\\
w/o depth selection & $\checkmark$ & -- & 57.6 & 7.0\\
w/o both & -- & -- & 56.9 & 7.0\\
\bottomrule
\end{tabularx}

\end{table}

\begin{table}[!t]
\caption{Complementary evidence selection ablation in SiLVR at fixed depth $K=7$. Both variants disable depth selection, and w/o both also removes MMR. Entries are accuracies (\%)}
\label{tab:rank_ablation}
\centering
\setlength{\tabcolsep}{4pt}
\renewcommand{\arraystretch}{1.10}
\begin{tabular*}{\columnwidth}{@{\extracolsep{\fill}}lcccc@{}}
\toprule
\textbf{Variant} & \textbf{Robot} & \textbf{Web} & \textbf{VideoMME-Long} & \textbf{Avg. Acc.}\\
\midrule
w/o depth selection & \textbf{41.1} & \textbf{56.7} & \textbf{57.6} & \textbf{51.8}\\
w/o both & 40.8 & 55.9 & 56.9 & 51.2\\
\bottomrule
\end{tabular*}

\end{table}

\begin{table}[t]
\caption{Query-guided pooling ablation in M3-Agent. The w/o pooling variant uses each clip's strongest query-matched node for relevance and representation. MMR and depth selection are retained}
\label{tab:internal_ablation}
\centering
\setlength{\tabcolsep}{3.8pt}
\renewcommand{\arraystretch}{1.08}
\begin{tabular*}{\columnwidth}{@{\extracolsep{\fill}}lcccc@{}}
\toprule
\multirow{2}{*}{\textbf{Benchmark}} & \multicolumn{2}{c}{\textbf{PACE-Mem}} & \multicolumn{2}{c}{\textbf{w/o pooling}}\\
\cmidrule(lr){2-3}\cmidrule(lr){4-5}
& Acc. (\%) & Clips & Acc. (\%) & Clips\\
\midrule
M3Bench Robot & \textbf{40.3} & 3.9 & 38.7 & 3.9\\
M3Bench Web & \textbf{57.6} & 4.1 & 56.7 & 4.1\\
VideoMME-Long & \textbf{58.7} & 4.3 & 57.6 & 4.3\\
\bottomrule
\end{tabular*}

\end{table}

\begingroup\setlength{\parskip}{0pt}
\resultsubsection{Main Results}
\label{sec:main_results}

\textbf{PACE-Mem achieves the best average accuracy.} It leads the evaluated methods in both host configurations (Table~\ref{tab:main}). These two configurations expose different memory interfaces: M3-Agent uses structured memories containing multiple records per clip, whereas SiLVR retrieves from a flat store of interval-level documents. Relative to Adaptive-$k$, the strongest baseline at comparable mean retrieval depths, PACE-Mem improves accuracy in all six evaluations, with an average increase of 1.6\%. Compared with Original M3-Agent, average accuracy rises from 44.2\% to 52.2\%.

\begingroup\clubpenalties 3 10000 10000 0
\textbf{Gains over fixed-depth retrieval span all three benchmarks.} At approximately four clips per call in M3-Agent, PACE-Mem improves over Fixed-4 by 1.9, 1.2, and 1.8 percentage points on Robot, Web, and VideoMME-Long, respectively. At approximately seven clips per call in SiLVR, the corresponding gains over Fixed-7 are 1.6, 1.3, and 2.1 percentage points. These consistent gains at similar mean depths indicate that evidence selection matters beyond the number of returned clips, across both multi-round and single-retrieval settings.\par\endgroup

\textbf{Larger fixed contexts do not close the average accuracy gap.} PACE-Mem achieves average accuracies of 52.2\% and 52.9\% with approximately 4.1 and 7.1 clips, compared with 51.1\% at 4.9 clips for Fixed-5 and 52.3\% at eight clips for Fixed-8. Increasing fixed depth improves baseline accuracy, but uniformly adding more relevance-ranked clips remains less effective on average than adapting evidence composition and depth to the question. Fixed-8 scores higher on Web (57.6\% versus 57.2\%), while PACE-Mem leads on Robot and VideoMME-Long.

\endgroup
\resultsubsection{Ablation Studies}
\label{sec:ablation}

Complementary ordering and depth selection both improve accuracy. We remove each selection component in SiLVR on VideoMME-Long (Table~\ref{tab:components}). Removing MMR applies the same depth rule to relevance order, while removing depth selection fixes the returned length at $K=7$. Accuracy decreases from 59.0\% for the full method to 58.1\% without MMR and 57.6\% without depth selection, and removing both yields 56.9\%. Depth selection improves accuracy by 1.2 percentage points under relevance ordering and 1.4 under MMR, showing that it benefits both ordering rules. Removing MMR increases mean returned depth from 7.2 to 7.4 even as accuracy falls, motivating a separate comparison of the ordering rules at a common depth.

Complementary ordering remains beneficial at fixed depth. Table~\ref{tab:rank_ablation} holds $K=7$ and disables depth selection across all three SiLVR benchmarks. Retaining MMR raises average accuracy from 51.2\% to 51.8\%, with gains of 0.3 percentage points on Robot, 0.8 on Web, and 0.7 on VideoMME-Long. With the clip count held fixed, these results show that changing evidence composition improves accuracy across the three benchmarks.

\begingroup\looseness=1
Query-guided pooling outperforms the strongest-node alternative. We replace pooled relevance and representations in M3-Agent with those of each clip's strongest query-matched node, retaining MMR and depth selection (Table~\ref{tab:internal_ablation}). Accuracy falls on all three benchmarks, with the average decreasing from 52.2\% to 51.0\% at the same reported mean depths. This result supports aggregating information beyond the strongest query match when constructing clip-level relevance scores and representations.\par\endgroup

\begin{figure*}[!t]
\begin{minipage}{\textwidth}
\centering
\setlength{\abovecaptionskip}{3pt}
\includegraphics[width=\textwidth,trim=0 4bp 0 0,clip]{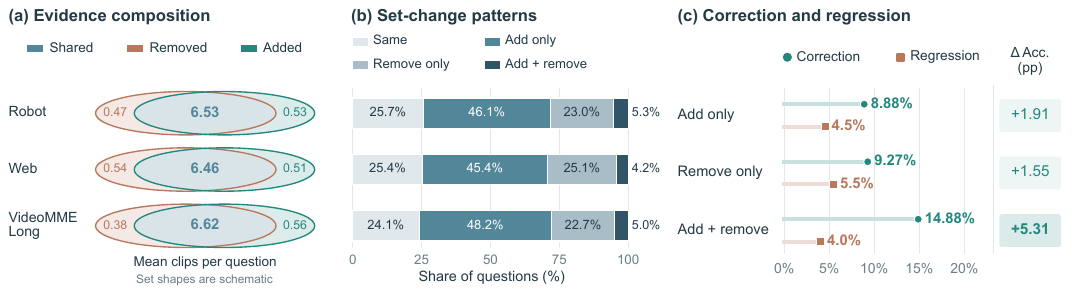}
\captionof{figure}{Evidence changes and answer outcomes in SiLVR relative to Fixed-7. (a) Mean clip counts in schematic sets. (b) Set-change frequencies; rows follow (a). (c) Rates pooled across benchmarks by change pattern. Correction and regression condition on initially wrong and initially correct answers, respectively. $\Delta$Acc. is corrections minus regressions divided by all questions in the group, in percentage points. Set comparisons disregard retrieval order.}
\label{fig:evidence_comparison}
\end{minipage}
\par\smallskip
\begin{minipage}{\textwidth}
\centering
\captionsetup[subfloat]{farskip=4pt}
\subfloat[]{\includegraphics[width=0.445\textwidth]{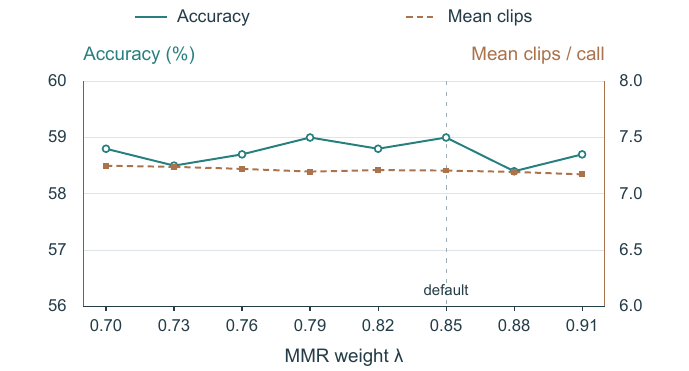}\label{fig:lambda_sensitivity}}\hfill
\subfloat[]{\includegraphics[width=0.525\textwidth]{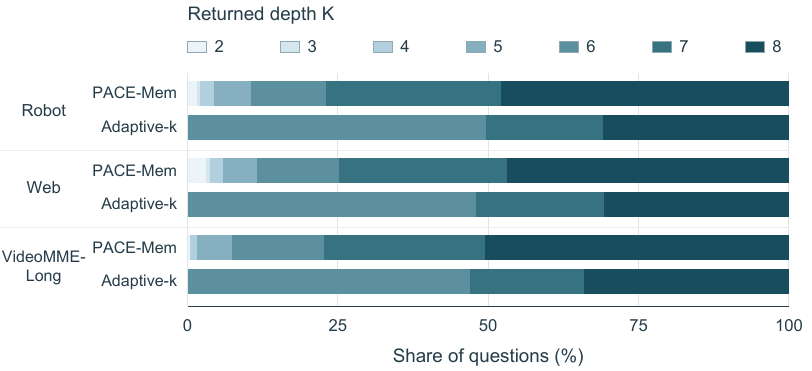}\label{fig:behavior}}
\captionof{figure}{Retrieval analysis in SiLVR. (a) Sensitivity to $\lambda$ on VideoMME-Long. Teal circles show accuracy (left axis), and copper squares show mean clip count (right axis). The dashed marker indicates $\lambda=0.85$. (b) Returned-depth distributions by benchmark and method. Adaptive-$k$ includes the five-clip buffer.}
\label{fig:retrieval_analysis}
\end{minipage}
\end{figure*}

\resultsubsection{Retrieval Analysis}
\label{sec:retrieval_analysis}

\begin{figure*}[!t]
  \centering
  \setlength{\abovecaptionskip}{0pt}
  \includegraphics[width=\textwidth]{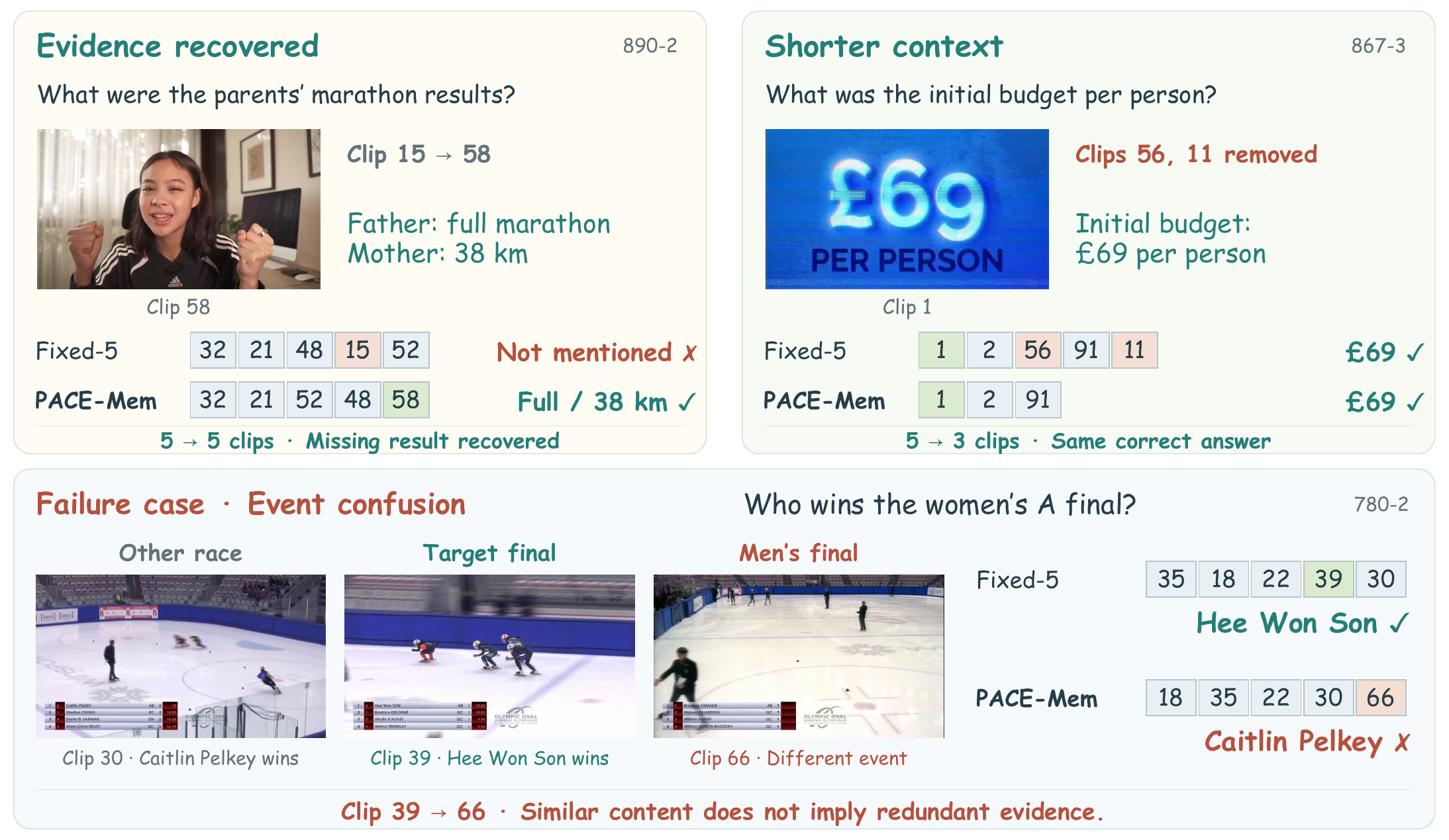}
  \caption{VideoMME-Long cases with M3-Agent under identical single search requests. Top left: Clip 58 recovers both parents' results at the same five-clip depth. Top right: retaining the explicit budget fact allows a correct answer with three clips instead of five. Bottom: omitting Clip 39 removes the target-final result, and PACE-Mem predicts the winner of another race. Frames and brief descriptions illustrate text-memory records. Rows show returned clip IDs and shortened predictions, with supporting evidence in green and replaced or omitted clips in peach.}
  \label{fig:cases}
\end{figure*}

\textbf{Evidence changes reveal the value of jointly adding and removing clips.} Fig.~\ref{fig:evidence_comparison} compares Fixed-7 and PACE-Mem in SiLVR's single-retrieval setting. Across Robot, Web, and VideoMME-Long, the methods share an average of 6.46--6.62 clips, while PACE-Mem adds 0.51--0.56 and removes 0.38--0.54. Thus, similar returned depths can contain different evidence: additions minus removals gives the net depth change, whereas their sum measures membership turnover. Add-only returns retain all Fixed-7 clips, remove-only returns are strict subsets, and mixed changes replace members through both operations. Add-only changes are most frequent, and identical sets account for approximately one quarter of questions (Fig.~\ref{fig:evidence_comparison}(b)). To relate these patterns to answer quality, Fig.~\ref{fig:evidence_comparison}(c) pools questions across benchmarks within each group. Correction and regression rates use initially wrong and initially correct Fixed-7 answers as their respective denominators; net accuracy gain divides corrections minus regressions by all questions in the group. Mixed changes have the largest within-group net gain, at 5.31 percentage points, compared with 1.91 for add-only and 1.55 for remove-only changes. They also have the highest correction rate (14.88\%) and lowest regression rate (4.0\%), although they occur in only 4.2--5.3\% of questions per benchmark. Removal-only changes correct a slightly larger fraction of errors than add-only changes, but also lose more initially correct answers, yielding a smaller net gain. One possible explanation for the mixed group's advantage is that replacement introduces missing cues while excluding repeated or competing event information. Adding alone retains potential distractions, whereas removing alone cannot supply a missing fact. This association makes evidence replacement a useful focus alongside context expansion and pruning. Across all 5,347 paired questions, PACE-Mem corrects 180 errors and introduces 100 regressions, yielding 80 additional correct answers; 2,707 questions are correct under both methods and 2,360 are incorrect under both. These pooled counts summarize question-level outcomes, while Table~\ref{tab:main} reports averages with equal benchmark weights.

\textbf{Performance is stable across the tested relevance weights.} We vary the MMR relevance weight $\lambda$ from $0.70$ to $0.91$ on SiLVR VideoMME-Long while keeping the other retrieval settings fixed. Larger values emphasize query relevance, while smaller values give more weight to avoiding overlap with selected clips. In Fig.~\ref{fig:retrieval_analysis}(a), accuracy varies nonmonotonically between 58.4\% and 59.0\%, and mean depth ranges from 7.17 to 7.25. The default $\lambda=0.85$ matches the highest accuracy at $\lambda=0.79$. The narrow accuracy and depth ranges indicate that the observed performance is maintained across several relevance--overlap balances within this interval, rather than depending on a single sharply tuned weight.

\textbf{PACE-Mem varies retrieval depth across questions.} Similar mean depths can conceal different evidence allocations, as shown in Fig.~\ref{fig:retrieval_analysis}(b). PACE-Mem returns two to eight clips, with fewer than eight returned for 52.2\% of Robot questions, 53.1\% of Web questions, and 49.4\% of VideoMME-Long questions. Eight is the most frequent depth and equals the configured cap $K_{\max}$. Prefixes of at most five clips account for 10.6\%, 11.6\%, and 7.3\% of questions, respectively, so shorter contexts serve a minority while most questions receive six to eight clips. Adaptive-$k$ mainly returns six to eight clips after its five-clip buffer, with six accounting for 47.0\%--49.6\% of questions. Adaptive-RAG returns seven clips for 804 of the 900 VideoMME-Long questions. Compared with these more concentrated allocations, PACE-Mem uses a broader range of depths while preserving longer contexts for most questions. Together with its higher average accuracy at similar mean depths (Table~\ref{tab:main}), this distribution supports adapting the returned prefix to the question-specific evidence trajectory rather than applying a nearly uniform context length.

\begingroup\setlength{\parskip}{0pt}\widowpenalty=10000
\resultsubsection{Case Studies}
\label{sec:examples}

Fig.~\ref{fig:cases} compares Fixed-5 and PACE-Mem on three VideoMME-Long questions under M3-Agent, illustrating recovered evidence, a shorter context, and event confusion. For each question, both methods issue the same single search request, and shared clips contain identical memory text. Clip IDs distinguish retained, omitted, and newly selected evidence.

\textbf{Replacing one clip recovers a missing result without increasing retrieval depth.} Question 890-2 asks whether the parents shown before the marathon participated and what results they achieved. Both methods return five clips and share four members. PACE-Mem replaces Clip 15, which describes support from the coach, companions, and parents without their race results, with Clip 58. The latter explicitly states that the father completed the full marathon and the mother reached 38 km. Fixed-5 predicts that the results were not mentioned, whereas PACE-Mem answers correctly. The replacement supplies the outcome asked for in the question, moving beyond related background about the parents' presence and support.

\setlength{\parskip}{0pt plus 0.2pt}
\textbf{Preserving an explicit answer fact allows a shorter context to retain the correct response.} Question 867-3 asks for the initial budget per person. Fixed-5 returns Clips 1, 2, 56, 91, and 11, while PACE-Mem returns the subset 1, 2, and 91. Both methods answer correctly because the retained context includes Clip 1's explicit \pounds69 budget. The omitted clips describe a later challenge (Clip 56) and an updated group budget (Clip 11), which also repeats the initial amount. Clip 11 therefore provides an additional mention of the answer, while Clip 1 alone already states the requested amount explicitly. The question's temporal qualifier distinguishes the requested initial amount from subsequent budget updates. The selected subset retains that direct support with three clips instead of five.

\textbf{Similar content does not necessarily provide redundant evidence.} Question 780-2 asks who wins the women's A final. Both methods return five clips and share four members, including Clip 30, which names Caitlin Pelkey as the winner of a different race. Fixed-5 also retains Clip 39, which explicitly identifies Hee Won Son as the winner of the target final, and answers correctly. PACE-Mem replaces Clip 39 with Clip 66, which introduces the men's 1000 m A final. It therefore loses the explicit result for the target race while retaining the competing winner from Clip 30, and incorrectly answers Caitlin Pelkey. Although these memories share race terminology and similar settings, they describe different events and their results are not interchangeable. This case illustrates why evidence selection must preserve facts that distinguish the target event from similar events, rather than treating topical similarity alone as evidence redundancy.


\setlength{\parskip}{0pt}
\compactsection{Conclusion}
\label{sec:conclusion}

We presented PACE-Mem, a training-free, plug-in retrieval method that jointly determines which memories to return and how much evidence to provide. A shared score trajectory connects complementary evidence selection with query-dependent retrieval depth. Across three benchmarks and two agent frameworks, PACE-Mem improves answer accuracy by an average of 1.6\% over Adaptive-$k$, the strongest baseline at comparable mean retrieval depths, without additional foundation-model calls for evidence selection. Results across structured and flat memories, together with fixed-depth ablations, support the importance of evidence composition alongside context length. Similar memories can nevertheless contain distinct event-specific facts required for an answer. Future work will explore preserving these distinctions during redundancy assessment and extending prefix selection to account for evidence accumulated across retrieval rounds.

\let\section\compactsection
\bibliographystyle{IEEEtran}
\bibliography{references}

@inproceedings{carbonell1998mmr,
  author    = {Jaime G. Carbonell and Jade Goldstein},
  title     = {The Use of {MMR}, Diversity-Based Reranking for Reordering Documents and Producing Summaries},
  booktitle = {Proc. ACM SIGIR Conf. Res. Develop. Inf. Retr.},
  pages     = {335--336},
  year      = {1998},
  doi       = {10.1145/290941.291025}
}

@inproceedings{lewis2020rag,
  author    = {Patrick Lewis and Ethan Perez and Aleksandra Piktus and Fabio Petroni and Vladimir Karpukhin and Naman Goyal and Heinrich K{\"u}ttler and Mike Lewis and Wen-tau Yih and Tim Rockt{\"a}schel and Sebastian Riedel and Douwe Kiela},
  title     = {Retrieval-Augmented Generation for Knowledge-Intensive {NLP} Tasks},
  booktitle = {Adv. Neural Inf. Process. Syst.},
  volume    = {33},
  pages     = {9459--9474},
  year      = {2020}
}

@inproceedings{jiang2023flare,
  author    = {Zhengbao Jiang and Frank F. Xu and Luyu Gao and Zhiqing Sun and Qian Liu and Jane Dwivedi-Yu and Yiming Yang and Jamie Callan and Graham Neubig},
  title     = {Active Retrieval Augmented Generation},
  booktitle = {Proc. Conf. Empirical Methods Natural Lang. Process. (EMNLP)},
  pages     = {7969--7992},
  year      = {2023},
  doi = {10.18653/v1/2023.emnlp-main.495}
}

@inproceedings{asai2024selfrag,
  author    = {Akari Asai and Zeqiu Wu and Yizhong Wang and Avirup Sil and Hannaneh Hajishirzi},
  title     = {Self-{RAG}: Learning to Retrieve, Generate, and Critique through Self-Reflection},
  booktitle = {Proc. Int. Conf. Learn. Represent. (ICLR)},
  pages     = {9112--9141},
  year      = {2024}
}

@inproceedings{jeong2024adaptiverag,
  author    = {Soyeong Jeong and Jinheon Baek and Sukmin Cho and Sung Ju Hwang and Jong C. Park},
  title     = {Adaptive-{RAG}: Learning to Adapt Retrieval-Augmented Large Language Models through Question Complexity},
  booktitle = {Proc. Conf. North Amer. Chapter Assoc. Comput. Linguistics (NAACL)},
  pages     = {7036--7050},
  year      = {2024},
  doi = {10.18653/v1/2024.naacl-long.389}
}

@inproceedings{taguchi2025adaptivek,
  author    = {Chihiro Taguchi and Seiji Maekawa and Nikita Bhutani},
  title     = {Efficient Context Selection for Long-Context {QA}: No Tuning, No Iteration, Just Adaptive-$k$},
  booktitle = {Proc. Conf. Empirical Methods Natural Lang. Process. (EMNLP)},
  pages     = {20105--20130},
  year      = {2025},
  doi       = {10.18653/v1/2025.emnlp-main.1017},
  url       = {https://aclanthology.org/2025.emnlp-main.1017/}
}

@misc{yan2024crag,
  author  = {Shi-Qi Yan and Jia-Chen Gu and Yun Zhu and Zhen-Hua Ling},
  title   = {Corrective Retrieval Augmented Generation},
  note    = {arXiv:2401.15884},
  year    = {2024}
}

@inproceedings{gutierrez2024hipporag,
  author    = {Bernal Jim{\'e}nez Guti{\'e}rrez and Yiheng Shu and Yu Gu and Michihiro Yasunaga and Yu Su},
  title     = {{HippoRAG}: Neurobiologically Inspired Long-Term Memory for Large Language Models},
  booktitle = {Adv. Neural Inf. Process. Syst.},
  volume    = {37},
  pages     = {59532--59569},
  year      = {2024},
  doi = {10.52202/079017-1902}
}

@misc{hu2026xmemory,
  author  = {Zhanghao Hu and Qinglin Zhu and Runcong Zhao and Di Liang and Hanqi Yan and Yulan He and Lin Gui},
  title   = {Beyond {RAG} for Agent Memory: Retrieval by Decoupling and Aggregation},
  note    = {arXiv:2602.02007v4},
  year    = {2026}
}

@inproceedings{long2026m3agent,
  author    = {Lin Long and Yichen He and Wentao Ye and Yiyuan Pan and Yuan Lin and Hang Li and Junbo Zhao and Wei Li},
  title     = {Seeing, Listening, Remembering, and Reasoning: A Multimodal Agent with Long-Term Memory},
  booktitle = {Proc. Int. Conf. Learn. Represent. (ICLR)},
  year      = {2026}
}

@article{zhang2026silvr,
  author  = {Ce Zhang and Yan-Bo Lin and Ziyang Wang and Mohit Bansal and Gedas Bertasius},
  title   = {{SiLVR}: A Simple Language-Based Video Reasoning Framework},
  journal = {Trans. Mach. Learn. Res.},
  year    = {2026}
}

@inproceedings{zhang2024llovi,
  author    = {Ce Zhang and Taixi Lu and Md Mohaiminul Islam and Ziyang Wang and Shoubin Yu and Mohit Bansal and Gedas Bertasius},
  title     = {A Simple {LLM} Framework for Long-Range Video Question-Answering},
  booktitle = {Proc. Conf. Empirical Methods Natural Lang. Process. (EMNLP)},
  pages     = {21715--21737},
  year      = {2024},
  doi = {10.18653/v1/2024.emnlp-main.1209}
}

@article{bai2024eventgraph,
  author  = {Ziyi Bai and Ruiping Wang and Difei Gao and Xilin Chen},
  title   = {Event Graph Guided Compositional Spatial--Temporal Reasoning for Video Question Answering},
  journal = {IEEE Trans. Image Process.},
  volume  = {33},
  pages   = {1109--1121},
  year    = {2024},
  doi     = {10.1109/TIP.2024.3358726}
}

@inproceedings{song2024moviechat,
  author    = {Enxin Song and Wenhao Chai and Guanhong Wang and Yucheng Zhang and Haoyang Zhou and Feiyang Wu and Haozhe Chi and Xun Guo and Tian Ye and Yanting Zhang and Yan Lu and Jenq-Neng Hwang and Gaoang Wang},
  title     = {{MovieChat}: From Dense Token to Sparse Memory for Long Video Understanding},
  booktitle = {Proc. IEEE/CVF Conf. Comput. Vis. Pattern Recognit. (CVPR)},
  pages     = {18221--18232},
  year      = {2024}
}

@inproceedings{shen2025longvu,
  author    = {Xiaoqian Shen and Yunyang Xiong and Changsheng Zhao and Lemeng Wu and Jun Chen and Chenchen Zhu and Zechun Liu and Fanyi Xiao and Balakrishnan Varadarajan and Florian Bordes and Zhuang Liu and Hu Xu and Hyunwoo J. Kim and Bilge Soran and Raghuraman Krishnamoorthi and Mohamed Elhoseiny and Vikas Chandra},
  title     = {{LongVU}: Spatiotemporal Adaptive Compression for Long Video-Language Understanding},
  booktitle = {Proc. ICML},
  pages     = {54582--54599},
  year      = {2025}
}

@inproceedings{fu2025videomme,
  author    = {Chaoyou Fu and Yuhan Dai and Yongdong Luo and Lei Li and Shuhuai Ren and Renrui Zhang and Zihan Wang and Chenyu Zhou and Yunhang Shen and Mengdan Zhang and Peixian Chen and Yanwei Li and Shaohui Lin and Sirui Zhao and Ke Li and Tong Xu and Xiawu Zheng and Enhong Chen and Caifeng Shan and Ran He and Xing Sun},
  title     = {Video-{MME}: The First-Ever Comprehensive Evaluation Benchmark of Multi-Modal {LLMs} in Video Analysis},
  booktitle = {Proc. IEEE/CVF Conf. Comput. Vis. Pattern Recognit. (CVPR)},
  pages     = {24108--24118},
  year      = {2025},
  doi = {10.1109/CVPR52734.2025.02245}
}

@misc{openai2024gpt4o,
  author  = {{OpenAI}},
  title   = {{GPT-4o} System Card},
  note    = {arXiv:2410.21276},
  year    = {2024},
  url     = {https://openai.com/index/gpt-4o-system-card/}
}

@misc{bai2025qwen3vl,
  author  = {Shuai Bai and Yuxuan Cai and Ruizhe Chen and Keqin Chen and Xionghui Chen and Zesen Cheng and Lianghao Deng and Wei Ding and Chang Gao and Chunjiang Ge and Wenbin Ge and Zhifang Guo and Qidong Huang and Jie Huang and Fei Huang and Binyuan Hui and Shutong Jiang and Zhaohai Li and Mingsheng Li and others},
  title   = {{Qwen3-VL} Technical Report},
  note    = {arXiv:2511.21631},
  year    = {2025}
}

@misc{zhang2025qwen3embedding,
  author  = {Yanzhao Zhang and Mingxin Li and Dingkun Long and Xin Zhang and Huan Lin and Baosong Yang and Pengjun Xie and An Yang and Dayiheng Liu and Junyang Lin and Fei Huang and Jingren Zhou},
  title   = {{Qwen3} Embedding: Advancing Text Embedding and Reranking Through Foundation Models},
  note    = {arXiv:2506.05176},
  year    = {2025}
}

@inproceedings{zhang2025flashvstream,
  author    = {Haoji Zhang and Yiqin Wang and Yansong Tang and Yong Liu and Jiashi Feng and Xiaojie Jin},
  title     = {{Flash-VStream}: Efficient Real-Time Understanding for Long Video Streams},
  booktitle = {Proc. IEEE/CVF Int. Conf. Comput. Vis. (ICCV)},
  pages     = {21059--21069},
  year      = {2025},
  doi = {10.1109/ICCV51701.2025.01957}
}

@inproceedings{murray2018lengthbias,
  author    = {Kenton Murray and David Chiang},
  title     = {Correcting Length Bias in Neural Machine Translation},
  booktitle = {Proc. Conf. Mach. Transl. (WMT)},
  pages     = {212--223},
  year      = {2018},
  doi       = {10.18653/v1/W18-6322},
  url       = {https://aclanthology.org/W18-6322/}
}

@article{shroff2010precis,
  author  = {Nitesh Shroff and Pavan Turaga and Rama Chellappa},
  title   = {Video Pr{\'e}cis: Highlighting Diverse Aspects of Videos},
  journal = {IEEE Trans. Multimedia},
  volume  = {12},
  number  = {8},
  pages   = {853--868},
  month   = dec,
  year    = {2010},
  doi     = {10.1109/TMM.2010.2058795}
}

@article{cheng2024krst,
  author  = {Yi Cheng and Hehe Fan and Dongyun Lin and Ying Sun and Mohan Kankanhalli and Joo-Hwee Lim},
  title   = {Keyword-Aware Relative Spatio-Temporal Graph Networks for Video Question Answering},
  journal = {IEEE Trans. Multimedia},
  volume  = {26},
  pages   = {6131--6141},
  year    = {2024},
  doi     = {10.1109/TMM.2023.3345172}
}

@article{xu2026rvlm,
  author  = {Jiaqi Xu and Cuiling Lan and Wenxuan Xie and Xuejin Chen and Yan Lu},
  title   = {Long Video Understanding With Learnable Retrieval in Video-Language Models},
  journal = {IEEE Trans. Multimedia},
  volume  = {28},
  pages   = {5254--5261},
  year    = {2026},
  doi     = {10.1109/TMM.2026.3664996}
}

@inproceedings{khan2026dfrag,
  author    = {Saadat Hasan Khan and Spencer Hong and Jingyu Wu and Kevin Lybarger and Youbing Yin and Erin Babinsky and Daben Liu},
  title     = {{DF-RAG}: Query-Aware Diversity for Retrieval-Augmented Generation},
  booktitle = {Findings Assoc. Comput. Linguistics: EACL},
  pages     = {2873--2894},
  year      = {2026},
  doi       = {10.18653/v1/2026.findings-eacl.150}
}

@inproceedings{kratzwald2018adaptive,
  author    = {Bernhard Kratzwald and Stefan Feuerriegel},
  title     = {Adaptive Document Retrieval for Deep Question Answering},
  booktitle = {Proc. Conf. Empirical Methods Natural Lang. Process. (EMNLP)},
  pages     = {576--581},
  year      = {2018},
  doi       = {10.18653/v1/D18-1055}
}

@misc{li2026evirerank,
  author  = {Minghan Li and Eric Gaussier and Juntao Li and Guodong Zhou},
  title   = {{EviRerank}: Adaptive Evidence Construction for Long-Document {LLM} Reranking},
  note    = {arXiv:2411.06254v6},
  year    = {2026},
  url     = {https://arxiv.org/abs/2411.06254v6}
}

@misc{luo2025univid,
  author  = {Jiabin Luo and Junhui Lin and Zeyu Zhang and Biao Wu and Meng Fang and Ling Chen and Hao Tang},
  title   = {{UniVid}: The Open-Source Unified Video Model},
  note    = {arXiv:2509.24200v2},
  year    = {2025}
}

@inproceedings{lu2014microblog,
  author    = {Kuang Lu and Diego Roa and Hui Fang},
  title     = {Concept Based Tie-breaking and Maximal Marginal Relevance Retrieval in Microblog Retrieval},
  booktitle = {Proc. Text REtrieval Conf. (TREC)},
  year      = {2014},
  url       = {https://trec.nist.gov/pubs/trec23/papers/pro-udel_fang_microblog.pdf}
}

@misc{peng2025adagres,
  author  = {Chao Peng and Bin Wang and Zhilei Long and Jinfang Sheng},
  title   = {{AdaGReS}: Adaptive Greedy Context Selection via Redundancy-Aware Scoring for Token-Budgeted {RAG}},
  note    = {arXiv:2512.25052v1},
  year    = {2025},
  url     = {https://arxiv.org/abs/2512.25052v1}
}

@misc{guo2026adagate,
  author  = {Yilin Guo and Yinshan Wang and Yixuan Wang},
  title   = {{AdaGATE}: Adaptive Gap-Aware Token-Efficient Evidence Assembly for Multi-Hop Retrieval-Augmented Generation},
  note    = {arXiv:2605.05245v1},
  year    = {2026},
  url     = {https://arxiv.org/abs/2605.05245v1}
}

@article{panda2017multiview,
  author  = {Rameswar Panda and Amit K. Roy-Chowdhury},
  title   = {Multi-View Surveillance Video Summarization via Joint Embedding and Sparse Optimization},
  journal = {IEEE Trans. Multimedia},
  volume  = {19},
  number  = {9},
  pages   = {2010--2021},
  month   = sep,
  year    = {2017},
  doi     = {10.1109/TMM.2017.2708981}
}

@IEEEtranBSTCTL{IEEEauthorcontrol,
  CTLuse_url = {no},
  CTLuse_forced_etal = {yes},
  CTLmax_names_forced_etal = {6},
  CTLnames_show_etal = {1}
}

@inproceedings{karpukhin2020dpr,
  author    = {Vladimir Karpukhin and Barlas Oguz and Sewon Min and Patrick Lewis and Ledell Wu and Sergey Edunov and Danqi Chen and Wen-tau Yih},
  title     = {Dense Passage Retrieval for Open-Domain Question Answering},
  booktitle = {Proc. Conf. Empirical Methods Natural Lang. Process. (EMNLP)},
  pages     = {6769--6781},
  year      = {2020},
  doi = {10.18653/v1/2020.emnlp-main.550}
}

@inproceedings{khattab2020colbert,
  author    = {Omar Khattab and Matei Zaharia},
  title     = {{ColBERT}: Efficient and Effective Passage Search via Contextualized Late Interaction over {BERT}},
  booktitle = {Proc. ACM SIGIR Conf. Res. Develop. Inf. Retr.},
  pages     = {39--48},
  year      = {2020},
  doi = {10.1145/3397271.3401075}
}

@inproceedings{izacard2021fid,
  author    = {Gautier Izacard and Edouard Grave},
  title     = {Leveraging Passage Retrieval with Generative Models for Open Domain Question Answering},
  booktitle = {Proc. Conf. Eur. Chapter Assoc. Comput. Linguistics (EACL)},
  pages     = {874--880},
  year      = {2021},
  doi = {10.18653/v1/2021.eacl-main.74}
}

@article{izacard2022contriever,
  author  = {Gautier Izacard and Mathilde Caron and Lucas Hosseini and Sebastian Riedel and Piotr Bojanowski and Armand Joulin and Edouard Grave},
  title   = {Unsupervised Dense Information Retrieval with Contrastive Learning},
  journal = {Trans. Mach. Learn. Res.},
  year    = {2022}
}

@inproceedings{park2023generativeagents,
  author    = {Joon Sung Park and Joseph C. O'Brien and Carrie J. Cai and Meredith Ringel Morris and Percy Liang and Michael S. Bernstein},
  title     = {Generative Agents: Interactive Simulacra of Human Behavior},
  booktitle = {Proc. ACM Symp. User Interface Softw. Technol. (UIST)},
  pages     = {1--22},
  year      = {2023},
  doi = {10.1145/3586183.3606763}
}

@misc{packer2023memgpt,
  author  = {Charles Packer and Sarah Wooders and Kevin Lin and Vivian Fang and Shishir G. Patil and Ion Stoica and Joseph E. Gonzalez},
  title   = {{MemGPT}: Towards {LLMs} as Operating Systems},
  note    = {arXiv:2310.08560},
  year    = {2023}
}

@inproceedings{shinn2023reflexion,
  author    = {Noah Shinn and Federico Cassano and Ashwin Gopinath and Karthik Narasimhan and Shunyu Yao},
  title     = {Reflexion: Language Agents with Verbal Reinforcement Learning},
  booktitle = {Adv. Neural Inf. Process. Syst.},
  volume    = {36},
  pages     = {8634--8652},
  year      = {2023},
  doi = {10.52202/075280-0377}
}

@inproceedings{kang2025memoryos,
  author    = {Jiazheng Kang and Mingming Ji and Zhe Zhao and Ting Bai},
  title     = {Memory {OS} of {AI} Agent},
  booktitle = {Proc. Conf. Empirical Methods Natural Lang. Process. (EMNLP)},
  pages     = {25961--25970},
  year      = {2025},
  doi       = {10.18653/v1/2025.emnlp-main.1318},
  url       = {https://aclanthology.org/2025.emnlp-main.1318/}
}

@inproceedings{xu2025amem,
  author    = {Wujiang Xu and Zujie Liang and Kai Mei and Hang Gao and Juntao Tan and Yongfeng Zhang},
  title     = {{A-Mem}: Agentic Memory for {LLM} Agents},
  booktitle = {Adv. Neural Inf. Process. Syst.},
  volume    = {38},
  pages     = {17577--17604},
  year      = {2025},
  doi       = {10.52202/085713-0593},
  url       = {https://proceedings.neurips.cc/paper_files/paper/2025/hash/19909c36f51abc4856b4560aff3d36d6-Abstract-Conference.html}
}

@inproceedings{chhikara2025mem0,
  author    = {Prateek Chhikara and Dev Khant and Saket Aryan and Taranjeet Singh and Deshraj Yadav},
  title     = {{Mem0}: Building Production-Ready {AI} Agents with Scalable Long-Term Memory},
  booktitle = {Proc. Eur. Conf. Artif. Intell. (ECAI)},
  pages     = {2993--3000},
  year      = {2025},
  doi       = {10.3233/FAIA251160}
}

@inproceedings{zhong2024memorybank,
  author    = {Wanjun Zhong and Lianghong Guo and Qiqi Gao and He Ye and Yanlin Wang},
  title     = {{MemoryBank}: Enhancing Large Language Models with Long-Term Memory},
  booktitle = {Proc. AAAI Conf. Artif. Intell.},
  volume    = {38},
  number    = {17},
  pages     = {19724--19731},
  year      = {2024},
  doi = {10.1609/aaai.v38i17.29946}
}

@inproceedings{zhang2023videollama,
  author    = {Hang Zhang and Xin Li and Lidong Bing},
  title     = {Video-{LLaMA}: An Instruction-Tuned Audio-Visual Language Model for Video Understanding},
  booktitle = {Proc. EMNLP Syst. Demonstr.},
  pages     = {543--553},
  year      = {2023},
  doi = {10.18653/v1/2023.emnlp-demo.49}
}

@article{ding2022languagehierarchy,
  author  = {Xinpeng Ding and Nannan Wang and Shiwei Zhang and Ziyuan Huang and Xiaomeng Li and Mingqian Tang and Tongliang Liu and Xinbo Gao},
  title   = {Exploring Language Hierarchy for Video Grounding},
  journal = {IEEE Trans. Image Process.},
  volume  = {31},
  pages   = {4693--4706},
  year    = {2022},
  doi     = {10.1109/TIP.2022.3187288}
}

@article{wang2022crossmodalgraph,
  author  = {Hao Wang and Guosheng Lin and Steven C. H. Hoi and Chunyan Miao},
  title   = {Cross-Modal Graph With Meta Concepts for Video Captioning},
  journal = {IEEE Trans. Image Process.},
  volume  = {31},
  pages   = {5150--5162},
  year    = {2022},
  doi     = {10.1109/TIP.2022.3192709}
}

@article{yang2023conceptaware,
  author  = {Bang Yang and Meng Cao and Yuexian Zou},
  title   = {Concept-Aware Video Captioning: Describing Videos With Effective Prior Information},
  journal = {IEEE Trans. Image Process.},
  volume  = {32},
  pages   = {5366--5378},
  year    = {2023},
  doi     = {10.1109/TIP.2023.3307969}
}

@article{xie2025captionassisted,
  author  = {Peiyu Xie and Jinxing Li and Guangming Lu and Yong Xu and David Zhang},
  title   = {Caption Assisted Multimodal Large Language Model for Video Moment Retrieval},
  journal = {IEEE Trans. Image Process.},
  volume  = {34},
  pages   = {6755--6766},
  year    = {2025},
  doi     = {10.1109/TIP.2025.3620124}
}

@inproceedings{radford2021clip,
  author    = {Alec Radford and Jong Wook Kim and Chris Hallacy and Aditya Ramesh and Gabriel Goh and Sandhini Agarwal and Girish Sastry and Amanda Askell and Pamela Mishkin and Jack Clark and Gretchen Krueger and Ilya Sutskever},
  title     = {Learning Transferable Visual Models from Natural Language Supervision},
  booktitle = {Proc. ICML},
  pages     = {8748--8763},
  year      = {2021}
}

@inproceedings{li2023blip2,
  author    = {Junnan Li and Dongxu Li and Silvio Savarese and Steven Hoi},
  title     = {{BLIP-2}: Bootstrapping Language-Image Pre-Training with Frozen Image Encoders and Large Language Models},
  booktitle = {Proc. ICML},
  pages     = {19730--19742},
  year      = {2023}
}

@article{liu2024lostmiddle,
  author  = {Nelson F. Liu and Kevin Lin and John Hewitt and Ashwin Paranjape and Michele Bevilacqua and Fabio Petroni and Percy Liang},
  title   = {Lost in the Middle: How Language Models Use Long Contexts},
  journal = {Trans. Assoc. Comput. Linguistics},
  volume  = {12},
  pages   = {157--173},
  year    = {2024},
  doi = {10.1162/tacl_a_00638}
}

@misc{ren2026icmbench,
  author  = {Shidu Ren and Yunze Liu and Xing Liu and Chi-Hao Wu and Enmin Zhou and Junxiao Shen},
  title   = {{ICM-Bench}: Person-Level Identity Reasoning in Multimodal Agents with Long-Term Memory},
  note    = {arXiv:2609.04438},
  year    = {2026},
  url     = {https://arxiv.org/abs/2609.04438}
}

@article{qian2024locans,
  author  = {Tianwen Qian and Ran Cui and Jingjing Chen and Pai Peng and Xiaowei Guo and Yu-Gang Jiang},
  title   = {Locate Before Answering: Answer Guided Question Localization for Video Question Answering},
  journal = {IEEE Trans. Multimedia},
  volume  = {26},
  pages   = {4554--4563},
  year    = {2024},
  doi     = {10.1109/TMM.2023.3323878}
}

@article{wang2024hsshg,
  author  = {Ruomei Wang and Yuanmao Luo and Fuwei Zhang and Mingyang Liu and Xiaonan Luo},
  title   = {{HSSHG}: Heuristic Semantics-Constrained Spatio-Temporal Heterogeneous Graph for {VideoQA}},
  journal = {IEEE Trans. Multimedia},
  volume  = {26},
  pages   = {11176--11190},
  year    = {2024},
  doi     = {10.1109/TMM.2024.3443661}
}
\endgroup

\end{document}